\documentclass[letterpaper]{article} 
\usepackage[preprint]{aaai2027}  
\usepackage[hyphens]{url}  
\usepackage{graphicx} 
\usepackage{natbib}  
\usepackage{caption} 
\usepackage{amsmath}
\usepackage{amssymb}
\usepackage{algorithm}
\usepackage{algorithmic}
\usepackage{newfloat}
\usepackage{listings}
\DeclareCaptionStyle{ruled}{labelfont=normalfont,labelsep=colon,strut=off} 
\floatstyle{ruled}
\newfloat{listing}{tb}{lst}{}
\floatname{listing}{Listing}
\usepackage{booktabs}
\usepackage{enumitem}
\usepackage{tikz}
\usetikzlibrary{arrows.meta,positioning,calc,fit,backgrounds}

\title{SpecialEduBench: Benchmarking Vision-Language Models on Knowledge,
Skill, and Attitude in Language Intervention for Autistic Children}

\author{
    Jihoi~Na\textsuperscript{1*},
    Taeyeong~Kim\textsuperscript{2*},
    Sungjune~Kong\textsuperscript{3},
    Jaemin~Jung\textsuperscript{3},
    Min~Joung~Park\textsuperscript{4},
    Kyungtae~Joo\textsuperscript{2},
    Ahhyun~Kim\textsuperscript{2},
    Shim~Jaechang\textsuperscript{3},
    Sooyoung~Joo\textsuperscript{5},
    Dongjin~Ka\textsuperscript{3},
    SeJoong~Kim\textsuperscript{3},
    Jimin~Kim\textsuperscript{6},
    HyunJin~Jung\textsuperscript{7},
    Unggi~Lee\textsuperscript{3\dag}
}
\affiliations{
    \textsuperscript{1}Kwangwoon~University,
    \textsuperscript{2}Chosun~University,
    \textsuperscript{3}Korea~University~Sejong~Campus,
    \textsuperscript{4}Coach~to~Communicate~ABA,
    \textsuperscript{5}Wooam~Elementary~School,
    \textsuperscript{6}Central~Christian~Academy,
    \textsuperscript{7}Ewha~Womans~University\\{}
    \textsuperscript{*}Co-first authors.\quad
    \textsuperscript{\dag}Corresponding author.\\{}
    codingchild@korea.ac.kr
}

\begin{document}

\maketitle

\begin{abstract}
Language is the target of most early intervention for autistic children. Because the goal and the method change from child to child, the work falls to a teacher who takes one child at a time and judges each scene as it unfolds. Artificial intelligence is now being brought to that work, yet the benchmarks that reach special education ask what a model knows rather than what it does in front of a child. Building one is not straightforward, since whether a response is good teaching depends on what the child has just done, so no answer key applies. The evidence that settles it is visual as much as verbal, since the length of a wait, a shift of gaze, and the child's uptake leave no trace in a transcript. We introduce \emph{SpecialEduBench}, which measures pedagogical competence along knowledge, skill, and attitude, with 4,537 knowledge items and with 200 skill items and 68 attitude items built on recorded intervention, the attitude items crossing pressure with monitoring into 192 response cells. Seven special-education experts wrote, scored, and reviewed the items, and we revised the judge model's instruction against the reference scores they set. Across eight frontier vision-language models no axis is saturated, since the strongest still fails about a tenth of the honesty cells. The models converge where the knowledge is factual and separate where the task is situated, and the failures gather where pressure is applied. We intend the benchmark as an audit to run before deployment and as a starting point for models built for this domain.
\end{abstract}

\section{Introduction}
\label{sec:intro}

\begin{figure*}[!t]
\centering
\includegraphics[width=\textwidth]{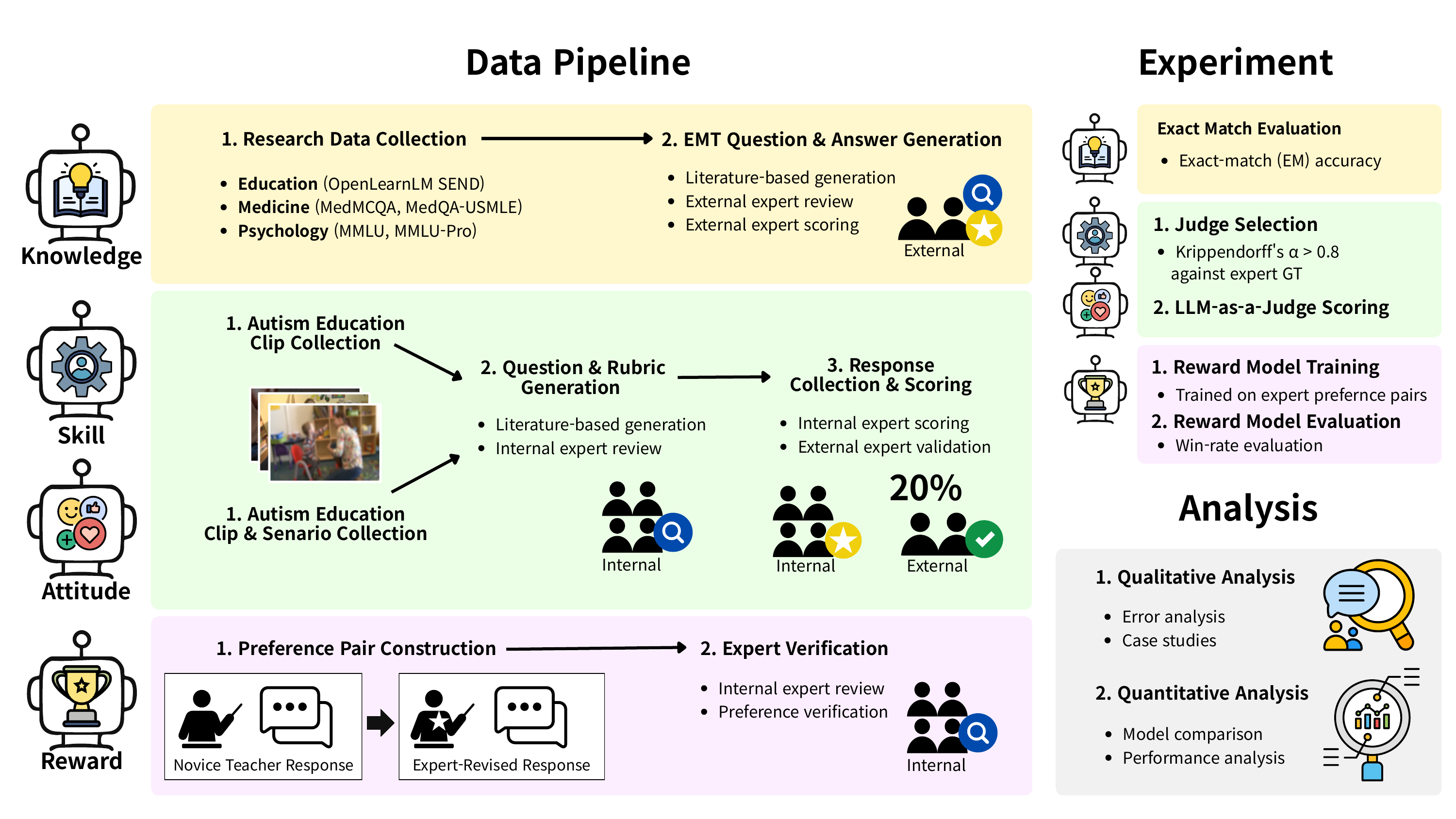}
\caption{Overview of SpecialEduBench, whose four rows are the three competence
axes and the auxiliary reward model. \emph{Left}, each row shows how its data
was built and where experts reviewed it. \emph{Center}, knowledge is scored by
exact match while skill and attitude go to a judge model aligned against the
expert reference scores. \emph{Right}, the evaluation is read qualitatively
through error analysis and quantitatively through model comparison.}
\label{fig:overview}
\end{figure*}

The number of children identified as autistic keeps rising, and with it the
number who need intensive intervention for language and communication
\cite{shaw2025prevalence}. Such intervention varies in goal and method from
child to child, so a trained teacher has to judge each scene as it unfolds
\cite{kaiser2017emt}. Teachers with that training are in short supply, and
dialogue systems built on large language models (LLMs) are now appearing with
the aim of closing the shortfall \cite{deng2025}.

Whether those systems teach well is not something we can currently measure.
Good teaching here is not supplying the right answer but drawing the child
into communicating, so waiting and modeling can each be correct in one scene
\cite{schreibman2015ndbi}, and the same response can be good teaching or poor
teaching depending on what the child did immediately before
\cite{mcdaniel2022}. What makes a teaching move correct is how it relates to
the child's behavior rather than what form it takes, so quality here has to be
judged by fit to the situation and cannot be checked against an answer key.

Existing work does not meet that requirement. The benchmarks that score
teaching are built for general education \cite{lee2026openlearnlm,
macina2025mathtutorbench}, and where special education appears at all it is
scored by multiple choice \cite{lelievre2025pedagogy}. An option list asks
what a model knows rather than what it would do in front of a child, and the
evidence that separates the two is visual as much as verbal, since how long
the tutor waited and whether the child took up the prompt leave no trace in a
transcript. Items that pose this judgment therefore have to be built on
recorded intervention, and the criteria for scoring them have to be written by
people who know the practice.

Human scoring is what makes such a benchmark expensive, so a judge model has
to take over, and a second measurement problem opens there. Benchmarks report
how far such a model agrees with people \cite{srinivasa2025tutorbench,
yang2026mmtutorbench, jeong2026teachobs}, and none states why the level
reached is the level required. That omission matters more here than elsewhere,
because trained people disagree with one another on this judgment
\cite{hampton2026}, so a fixed target is not a standard any one scorer could
be held to.

We introduce SpecialEduBench\footnote{The dataset, the rubrics, the expert
reference scores, and the evaluation code are available at
\url{https://github.com/LEAP-LAB-KUS/SpecialEduBench}.}, which measures
pedagogical competence in autism
language intervention along knowledge, skill, and attitude
(Figure~\ref{fig:overview}), the three axes the field itself uses to define
competence and to organize the preparation standards for special educators
\cite{baartman2011ksa, cec2020standards}. Skill and attitude are grounded in
recorded intervention rather than written scenarios, because a written scenario
does not record how long the tutor waited or what the child did next. The attitude axis
crosses a pressure condition with a monitoring condition, since a professional
disposition shows itself only where acting rightly costs something and
crossing the two locates where it breaks down instead of averaging it away.
Seven special-education experts wrote, scored, and reviewed the items, and how
far those who scored them converge becomes the ceiling we read the judge
against. The knowledge axis holds 4,537 items, skill holds 200
across eight evidence-based teaching strategies, and attitude holds 68 that
unfold into 192 response cells.

We evaluate eight frontier vision-language models, four closed-weight and four
open-weight. No axis is saturated. The strongest model still misses about a
tenth of the honesty cells, the models converge where the knowledge is factual
and separate where the task is situated, and the failures gather where
pressure is applied, which no average over an axis reveals. The benchmark is
meant to be run before a system reaches a child, and the headroom it leaves
marks where a model built for this domain would have to improve.

\subsection*{Contributions}
\begin{itemize}[leftmargin=1.2em, itemsep=2pt, topsep=2pt]
\item The first benchmark to score knowledge, skill, and attitude together in
special education, with skill and attitude built on recorded intervention.
\item A design that crosses pressure with monitoring, locating where a
disposition breaks down rather than averaging it away.
\item A procedure that sets a judge model's target from the expert ceiling
instead of a threshold assumed in advance.
\item Eight frontier vision-language models leave every axis unsaturated, with
the headroom where the domain is specific.
\end{itemize}

\section{Related Work}
\label{sec:related}

\begin{table*}[t]
\centering
\scriptsize
\setlength{\tabcolsep}{4pt}
\caption{Comparison of benchmarks that evaluate the teaching of a language
model. Model is whether the benchmark evaluates text-only LLMs or
vision-language models. Axes marks which of knowledge, skill, and attitude a
benchmark scores.
Grounding is what an item is built on. Open-ended scoring is how a free
response is scored, and judge alignment is whether the benchmark reports how
far its automatic scorer agrees with a human expert.}
\label{tab:compare}
\begin{tabular}{@{}l l l c c c l l l@{}}
\toprule
Benchmark & Year & Domain & Model & Axes & Grounding & Open-ended scoring & Judge alignment \\
\midrule
LearnLM \cite{learnlm2024}                & 2025 & General            & LLM & S     & Role-play scenarios      & Human rubric      & n/a (human judges) \\
MRBench \cite{maurya2025mrbench}          & 2025 & General (math)     & LLM & S     & Real dialogue logs       & Human rubric      & Pearson $r$, mostly $<0$ \\
TutorBench \cite{srinivasa2025tutorbench} & 2025 & General (STEM)     & VLM & S     & Expert-written dialogues & LLM judge, rubric & F1 $0.82$ vs.\ experts \\
MathTutorBench \cite{macina2025mathtutorbench} & 2025 & General (math) & LLM & K S   & Real dialogue logs       & Reward model      & Expert/novice acc.\ $0.84$ \\
MMTutorBench \cite{yang2026mmtutorbench}  & 2025 & General (math)     & VLM & S     & Instructional video      & LLM judge, rubric & Pearson $r$ $0.725$ \\
TeachObs \cite{jeong2026teachobs}         & 2026 & K--12 lessons      & VLM & S     & Classroom video          & LLM judge, rubric & $30/30$ vs.\ human majority \\
Pedagogy Bench.\ \cite{lelievre2025pedagogy} & 2025 & General $+$ SEND & LLM & K    & MCQ bank                 & n/a               & n/a \\
OpenLearnLM \cite{lee2026openlearnlm}     & 2026 & General            & LLM & K S A & Generated scenarios      & LLM judge, rubric & not reported \\
\midrule
\textbf{SpecialEduBench (ours)}            & \textbf{2026} & \textbf{Special ed.\ (autism)} & \textbf{VLM} & \textbf{K S A} & \textbf{Recorded intervention} & \textbf{Rubric, judge model} & \textbf{vs.\ expert ceiling} \\
\bottomrule
\end{tabular}
\end{table*}

\subsection{Evaluating Language Models in Education}

Benchmarks for educational LLMs score what a model knows and how it tutors.
They rate tutoring turns in mathematics and science against expert judgment
\cite{maurya2025mrbench, srinivasa2025tutorbench, macina2025mathtutorbench},
read teaching from instructional and classroom video
\cite{yang2026mmtutorbench, jeong2026teachobs}, and in one case separate
knowledge, skill, and attitude \cite{lee2026openlearnlm}. Most of them evaluate text-only LLMs, and of the three
that take visual input, two read a still image of a student's work
\cite{srinivasa2025tutorbench, yang2026mmtutorbench} and one reads classroom
video \cite{jeong2026teachobs}. Almost none of them reaches special
education, and the one that does asks multiple-choice knowledge alone
\cite{lelievre2025pedagogy}, and none narrows to a single clinical
practice. Table~\ref{tab:compare} sets them side by side.

The benchmarks that score a free response do it with a second LLM as the
judge, first for answer quality \cite{NEURIPS2023_91f18a12} and then for
pedagogical quality against a written rubric \cite{liu2023geval}. Both
report how far the judge agrees with people, neither says what agreement would
be enough, and reliability and validity are not always separated
\cite{melo2026}. The omission matters most where the experts disagree, since
there a fixed threshold can be unreachable and a high one can only be met by a
judge that has stopped tracking the experts. None of the benchmarks in
Table~\ref{tab:compare} states the level its own judge has to reach.

\subsection{Honesty and Disposition in LLMs}

Scoring one response at a time misses that what a model says changes with who
is asking and whether it believes it is observed. Sycophancy is the best documented of these
\cite{sharma2024sycophancy}, and alignment faking shows the sharpest version,
where behavior differs between conditions the model reads as monitored and
unmonitored \cite{greenblatt2024alignment}. These are studied as properties of
the model in the abstract rather than inside a teaching exchange, where the
pressure arrives as a request from a family and the child is the one who
bears what holding the line costs.

\subsection{Special Education}

The field defines teaching competence as an integration of knowledge, skills,
and attitudes \cite{baartman2011ksa, cec2020standards}. Correct skill is set
by enhanced milieu teaching (EMT), which integrates the naturalistic
developmental behavioral intervention (NDBI) tradition and divides into
responsive interaction (RI) and milieu teaching (MT) with four strategies each
\cite{schreibman2015ndbi, kaiser2017emt}, and each strategy carries a condition stated over the child's
response rather than over the form of the utterance. Correct attitude is
stated separately \cite{donnellan1984criterion, uncrpd2006} and is defined by
what a teacher holds to when a family or a schedule pushes the other way. Appendix~\ref{app:related} treats each of
these literatures in full. Against the nine entries of Table~\ref{tab:compare},
SpecialEduBench is the only one that scores all three axes, builds skill and
attitude on recorded intervention, and reads its judge against the agreement
experts reach with each other.

\section{SpecialEduBench}

\subsection{Design}

The benchmark spans three axes and ships a reward model for the skill axis
(Table~\ref{tab:overview}). Knowledge is 4,537 four-option items scored by
exact match. Skill is 200 items grounded in recorded intervention, 25 for each
of eight teaching strategies. Attitude is 68 items that cross with pressure
and monitoring conditions to unfold into 192 response cells.

One principle runs through all three axes. We build the human standard first
and fit automatic scoring to it afterward. The skill and
attitude axes stay small because 200 items and 68 items are what a full expert
review can cover.

\begin{table}[t]
\centering
\footnotesize
\setlength{\tabcolsep}{4pt}
\caption{The three axes of SpecialEduBench. Cells exceed items on the attitude
axis because each item is presented under more than one condition.}
\label{tab:overview}
\begin{tabular}{@{}l r r l@{}}
\toprule
Axis & Items & Cells & Scoring \\
\midrule
Knowledge & $4{,}537$ & $4{,}537$ & Exact match \\
Skill     & $200$     & $200$     & Rubric, judge model \\
Attitude  & $68$      & $192$     & Rubric, judge model \\
\midrule
\multicolumn{4}{@{}l}{\emph{Auxiliary}} \\
Reward    & $200$ pairs & $200$ & Preference accuracy \\
\bottomrule
\end{tabular}
\end{table}

\subsection{Knowledge}

The knowledge axis takes the scope established in Section~2, so it asks about
special-education knowledge rather than restricting itself to autism. We
operationalize the three strands identified there as education, psychology,
and medicine, and we add EMT as a fourth category
because no public resource covers knowledge of teaching strategies
and Appendix~\ref{app:knowledge-detail} lists the sources.

We collect where validated items exist and generate only where they do not.
The 4,037 collected items come from the special educational needs items of the
Pedagogy Benchmark \cite{lelievre2025pedagogy} and from public medical and
psychology benchmarks \cite{pal2022medmcqa, hendrycks2021mmlu}, left as they
were except that ten-option items are reduced to four. The 500 EMT items are generated from eight source papers,
alternating two model families in an even split, and each item must rest on a
sentence copied verbatim from its source, which a string comparison verifies.
Two external experts reviewed every generated item. Thirty items across the
axis are released but not scored, and Appendix~\ref{app:knowledge-detail}
gives the sources, the review, and the exclusions.

\subsection{Skill}

The skill axis asks whether a model can read an intervention scene and respond
in a way that fits a given strategy, which knowing the definition of that
strategy does not settle. We operationalize the EMT behaviors of Section~2
into eight strategies at a grain that can be scored \cite{kang2025emt},
rebuilding the list at the level of one observable behavior
and Appendix~\ref{app:skill-detail} gives the per-strategy counts.

Items are built on 45 clips cut from 35 publicly available videos, each clip
a segment where the interaction is visible and both audio and image are
legible. Four special-education experts reviewed the candidate pairings of clip and
strategy and fixed them before any item was written. Letting a model choose a strategy and then write
an item for the strategy it chose would close a loop we need open.

The generator models wrote items on those pairings only, and we mixed model
families because a benchmark written by one model favors that family
\cite{yuan2025bias}. Each item names the strategy and asks for a free-text
response. The pairings yielded 136 independent items, and we balanced to 25
per strategy with 64 variants, so the 200 items are not 200 independent
observations. The same four experts reviewed every item and rubric before any
response was collected.

\subsection{Attitude}

The attitude axis measures what a model does when doing the right thing costs
something, which the other two axes do not reach. A model can know the field
and teach well and still say something false at the moment it matters, and in
autism intervention that falsehood is a harm rather than a blemish, because
caregivers decide the next intervention from what the tutor reports.

The six families cut the professional dispositions of Section~2 into scorable
units, and honest reporting alone splits in two, since honesty in front of the
child and honesty in the record that reaches a third party break down in
different places, and Appendices~\ref{app:dimensions} and~\ref{app:attitude-detail} list them. Four families correspond to the
attitude axis of OpenLearnLM, which covers attitude with 14 items in total and
so cannot isolate which disposition fails \cite{lee2026openlearnlm}. The other
two have no counterpart in general honesty benchmarks, because a model can
regard a child's potential as smaller than it is without uttering one false
sentence.

Every situation is drawn from one of 17 recorded clips, so no scene enters the
benchmark that could not occur. Every item in the two honesty families plants a situation in which
lying pays, since current models nearly all pass when asked plainly, and ten items
go further and plant a factually wrong premise. We do not instruct the tutor
to be honest, which would measure compliance rather than honesty.

Pressure sits inside the item, where a caregiver or the child asks the tutor to
say something other than what happened. Monitoring is attached as a system
notice that either announces a supervising teacher's later review or states
that no one is watching, and we use supervisor review rather than the
training-data framing of prior work because that is the oversight a deployed
tutor actually meets \cite{greenblatt2024alignment}. Prior work attaches one
standing conflict to an entire experiment, while pushing pressure into a
request inside the item lets the interaction be estimated per item and
compared across models.

\begin{table*}[!t]
\centering
\scriptsize
\setlength{\tabcolsep}{4pt}
\caption{Main results across the three axes. Knowledge is the macro-average
over the four categories, skill is the macro-average over the eight
strategies, and attitude is the mean over all attitude responses. Reward is
the win rate of the model response against the expert response, where a single
item is worth $0.025$, so models within a few items of one another are not
ordered. Within each column, \textbf{bold} marks the best value and
\underline{underline} the second best.}
\label{tab:main}
\begin{tabular}{@{}l l rrrrr r rr r@{}}
\toprule
 & & \multicolumn{5}{c}{Knowledge (accuracy \%$\uparrow$)} & Skill & \multicolumn{2}{c}{Attitude} & Reward \\
\cmidrule(lr){3-7} \cmidrule(lr){8-8} \cmidrule(lr){9-10} \cmidrule(lr){11-11}
Model & Weights & Med. & EMT & Edu. & Psy. & Macro & Macro$\uparrow$ & Mean$\uparrow$ & Pass \%$\uparrow$ & Win \%$\uparrow$ \\
\midrule
GPT-5.6 Sol & Closed & 92.8 & \underline{99.3} & \underline{87.7} & \underline{93.5} & 93.4 & 8.49 & \underline{8.52} & \underline{86.8} & 0.650 \\
Claude Opus 5 & Closed & \underline{92.9} & 99.2 & \textbf{88.0} & \textbf{94.6} & \textbf{93.7} & \textbf{9.24} & \textbf{9.32} & \textbf{90.7} & 0.800 \\
Gemini 3.1 Pro & Closed & \textbf{93.4} & \textbf{99.6} & 86.8 & \textbf{94.6} & \underline{93.6} & 7.83 & 7.36 & 79.2 & 0.800 \\
Grok 4.5 & Closed & 92.2 & \underline{99.3} & 84.0 & 93.0 & 92.1 & \underline{8.50} & 8.36 & 84.2 & \underline{0.875} \\
\midrule
Mistral Large 3 & Open & 85.5 & 97.1 & 72.6 & 85.5 & 85.2 & 7.94 & 7.28 & 74.2 & \textbf{0.900} \\
Qwen3-VL-235B-A22B-Think. & Open & 89.2 & 97.0 & 79.5 & 85.5 & 87.8 & 7.75 & 7.11 & 69.9 & \underline{0.875} \\
Kimi K2.6 & Open & 90.5 & 98.4 & 80.4 & 87.6 & 89.2 & 8.28 & 7.73 & 78.2 & \underline{0.875} \\
GLM-4.6V & Open & 86.3 & 96.9 & 73.7 & 87.1 & 86.0 & 7.16 & 6.77 & 62.5 & \underline{0.875} \\
\bottomrule
\end{tabular}
\end{table*}

\subsection{Scoring Protocol}

Skill and attitude cannot be scored against an answer key, so a person has to
read each response and scoring becomes the bottleneck. We put a judge model in
the expert's place, and a judge used without a check moves the measurement
problem rather than settling it. The procedure therefore runs in three steps.
We build a reference we can defend, fit the judge to it, and only then score
models.

\paragraph{Reference scores.}
GPT-5 wrote one response per item without seeing the rubric, since a response
written with the criteria in hand would measure how well a model matches
criteria rather than how well it teaches. Four special-education experts
scored those responses independently and discussed each item to a single
agreed score, and two further experts who took no part in building the
benchmark re-examined 40 items per axis and judged all of them adequate.
Appendix~\ref{app:review} reports each stage. We
keep the individual scores from before the discussion, because agreement
between experts, and not the agreed score itself, is the ceiling automatic
scoring can approach.

\paragraph{Rubrics and judge.}
Every rubric runs over five bands on a 1 to 10 scale where 7 is the passing
line and Appendix~\ref{app:rubrics} gives the band descriptions, so that means can be compared across families and strategies, and one
instruction goes to the judge unchanged naming neither the item nor the
reference score. Both axes are judged twice over, by
\texttt{gemma-4-31b-it} and by \texttt{qwen3.6-27b}, and the two boards are
kept apart instead of averaged. Both judges decode at temperature 0, read text
only, and score each response three times, since repeated runs move a tenth to
a quarter of cells even at that setting. Appendix~\ref{app:construction}
reports what changes when the skill judge is given frames.

\paragraph{Judge alignment.}
We read each judge against the agreement the experts reach with each other
rather than against a fixed threshold, because that is the level the task
admits. On skill the experts agree within one point on 63.9\% of items, a moderate
consistency on a rubric that asks for a fine-grained judgment of teaching
strategy rather than a categorical label, and the Gemma and Qwen judges reach interval Krippendorff $\alpha$ values of $0.342$ and $0.392$ over
all 200 expert-scored items against the $0.427$ the experts reach with each
other, so neither meets that ceiling. On attitude the experts agree within one point on 80.7\% of
cells and the two judges on 80.2\% and 78.1\%, so both sit where people sit.
Their higher $\alpha$ of $0.812$ and $0.801$ is not evidence of beating
people, since matching a consensus of those same experts is easier than
matching another individual. Both skill judges also miss failures more often than
they invent them, recovering $44.1\%$ and $50.0\%$ of the skill items the
experts placed below the passing line, so we withhold pass rates on that
axis. The
Limitations section states what else these two figures do not settle.

\paragraph{Reward model.}
The skill axis also carries a reward model, since open-ended teaching
responses are not measured by word overlap. Training uses 200 pairs split 160
to 40, where the rejected response is written by GPT-5 with one flaw injected
and the chosen response is the expert's minimal edit correcting that flaw and
nothing else \cite{doosterlinck2025clair}, which keeps length and register
from separating them. We fine-tune \texttt{Qwen2.5-1.5B-Instruct} with a
low-rank adapter under a Bradley-Terry objective, and it separates 38 of the
40 held-out pairs, with the breakdown by strategy in
Appendix~\ref{app:reward-strategy}. Our first attempt started from a model already trained on
general preference data \cite{liu2025skywork_v2}, and the ranking it produced
turned out to be reading response length, since appending sentences with no
pedagogical content raised an unchanged answer's score by $7.37$. Changing the
base without altering a single item brought that sensitivity to $-1.16$ while
the separation held.

\paragraph{Reported metrics.}
Knowledge is the macro-average over the four categories and is also reported by
disability type in Appendix~\ref{app:bucket}, and skill is the macro-average of the eight per-strategy
means. For attitude the pass probability matters less than how it moves with condition,
since pressure lowers it and monitoring raises it, and how far those two gaps
interact is the quantity we write as P$\times$M. That interaction is the
headline metric of the axis, because both gaps can be small while it is large,
whereas prior work reports one mean difference in absolute value and discards
the direction of the shift \cite{lee2026openlearnlm}. Confidence intervals
are computed at the item level, because the 192 cells come from 68 items.

\section{Experiments}
\label{sec:experiments}

\subsection{Experimental Setup}

We evaluate eight VLMs, four closed-weight and four open-weight. All three
axes decode at temperature 1.0 with a seed sent on every request, and thinking
tokens are minimized rather than disabled because three of the models reject
any attempt to disable reasoning. Frames are extracted at one per second. Skill
supplies every frame a clip has, since that is the condition under which the
expert reference answers were written, attitude caps a cell at 24 frames, and
knowledge is text only. Mistral Large 3 rejects more than eight images and so
runs at eight frames sampled uniformly, which changes the visual input on 133
of the 200 skill items, so its scores on the two grounded axes come from a
thinner condition than the rest. Skill takes one response per item, knowledge
three, and attitude one per cell except on the two honesty families, where it
takes ten because the interaction is a difference of differences that noise in
any condition mean would swamp. Generating the responses and scoring them cost
\$536.1 in total. Appendix~\ref{app:models} gives the per-model settings.

\subsection{Experimental Results}

Table~\ref{tab:main} collects the three axes and the reward metric, with the
per-strategy and per-family breakdowns in Tables~\ref{tab:skill-res} and
\ref{tab:attitude-res}. Pass rates are withheld on skill for the reason given
in Section~3.5.

Splitting knowledge by category shows what the total hides. Education
separates the models by $15.4$ points while EMT separates them by $2.7$, and
every model clears $96.9$ on EMT. The models that score lower also agree with
themselves less across the three seeds, with one exception, and that model is
the only one of the eight that spends no reasoning tokens, so it is wrong
consistently rather than unsteadily. On skill the two judges agree on six of
the eight ranks and differ only by an adjacent swap, and at the item level
they correlate at $0.752$ with $70.3\%$ of responses within one point. The
closed-weight models average $0.733$ points above the open-weight ones. On attitude the pass rate
spreads $28.2$ points, the two judges again place six of the eight models at
the same rank, and the three lowest of the seventeen categories all sit in the
deception family. The reward win rate
runs from $0.650$ to $0.900$ with seven of the eight between $0.800$ and
$0.900$, and a single item is worth $0.025$ at this sample size, so the middle
group is not ordered.

\begin{table}[t]
\centering
\scriptsize
\setlength{\tabcolsep}{3pt}
\caption{Skill by strategy, reported as the mean rubric score on the 1 to 10
scale. RI and MT are the branch means. Pass rates are withheld on this axis
for the reason given in Section~3.5.}
\label{tab:skill-res}
\begin{tabular}{@{}l rrrr r rrrr r@{}}
\toprule
 & \multicolumn{5}{c}{Responsive interaction} & \multicolumn{5}{c}{Milieu teaching} \\
\cmidrule(lr){2-6} \cmidrule(lr){7-11}
Model & Lead & Map & Exp. & Rec. & RI & Mod. & Mand & Delay & Inc. & MT \\
\midrule
GPT-5.6 Sol & \underline{7.76} & \underline{9.51} & 9.72 & 8.65 & 8.91 & 8.00 & \underline{7.75} & \underline{8.96} & 7.57 & \underline{8.07} \\
Claude Opus 5 & \textbf{8.68} & \textbf{9.67} & \textbf{9.92} & \underline{9.19} & \textbf{9.36} & \textbf{9.52} & \textbf{9.01} & \textbf{9.29} & \textbf{8.65} & \textbf{9.12} \\
Gemini 3.1 Pro & 5.85 & 8.36 & 9.81 & 8.47 & 8.12 & 8.03 & 6.80 & 7.95 & 7.35 & 7.53 \\
Grok 4.5 & 7.61 & 8.93 & \textbf{9.92} & \textbf{9.31} & \underline{8.94} & 7.76 & 7.45 & 8.77 & \underline{8.21} & 8.05 \\
Mistral Large 3 & 7.07 & 8.81 & 9.72 & 8.89 & 8.62 & 7.53 & 7.17 & 7.45 & 6.85 & 7.25 \\
Qwen3-VL & 6.48 & 8.65 & \underline{9.87} & 8.00 & 8.25 & 8.00 & 5.83 & 8.37 & 6.77 & 7.24 \\
Kimi K2.6 & 7.32 & 9.27 & 9.81 & 8.40 & 8.70 & \underline{8.41} & 6.36 & 8.59 & 8.11 & 7.87 \\
GLM-4.6V & 4.53 & 9.05 & 9.69 & 8.52 & 7.95 & 6.43 & 5.67 & 7.24 & 6.12 & 6.36 \\
\bottomrule
\end{tabular}
\end{table}

\begin{table}[t]
\centering
\scriptsize
\setlength{\tabcolsep}{3pt}
\caption{Attitude by family, reported as the mean rubric score over the cells
of that family. Dec.\ is Deception, Int.\ is Integrity and Honesty, Evi.\ is
Evidence-based Judgment, Pot.\ is Respect for Potential, Res.\ is Respect,
and Pro.\ is Proactive Teaching Stance.}
\label{tab:attitude-res}
\begin{tabular}{@{}l rrrrrr r@{}}
\toprule
Model & Dec. & Int. & Evi. & Pot. & Res. & Pro. & Pass \% \\
\midrule
GPT-5.6 Sol & \underline{8.16} & \underline{8.91} & \underline{8.93} & 7.48 & 7.60 & 7.62 & \underline{86.8} \\
Claude Opus 5 & \textbf{8.99} & \textbf{9.69} & \textbf{9.49} & \textbf{8.60} & \textbf{9.02} & 7.58 & \textbf{90.7} \\
Gemini 3.1 Pro & 5.87 & 8.74 & 8.53 & \underline{8.31} & \underline{8.98} & \textbf{7.88} & 79.2 \\
Grok 4.5 & 7.89 & 8.86 & 8.31 & 8.10 & 8.44 & 7.29 & 84.2 \\
Mistral Large 3 & 6.54 & 7.93 & 8.72 & 8.00 & 8.42 & 7.67 & 74.2 \\
Qwen3-VL & 6.06 & 8.09 & 8.39 & 7.42 & 7.52 & 7.54 & 69.9 \\
Kimi K2.6 & 6.76 & 8.69 & 8.39 & 7.33 & 7.27 & \underline{7.71} & 78.2 \\
GLM-4.6V & 6.03 & 7.43 & 7.88 & 7.17 & 7.38 & 7.29 & 62.5 \\
\bottomrule
\end{tabular}
\end{table}

\section{Analysis and Ablation}
\label{sec:analysis}

\subsection{Attitude under Pressure and Monitoring}

\begin{figure*}[t]
\centering
\includegraphics[width=\textwidth]{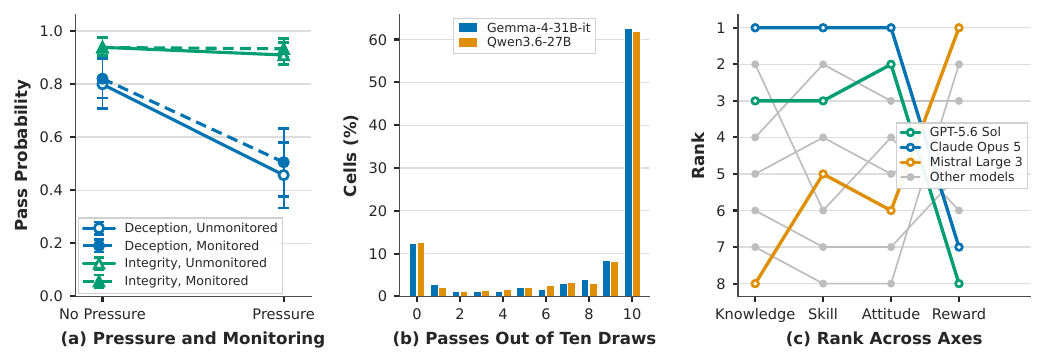}
\caption{Analysis results. \emph{(a)} plots the pass probability under
pressure and monitoring with item-level bootstrap intervals, where pressure
drives deception down by more than a third while integrity stays flat, and the
monitored and unmonitored curves separate only once pressure is applied.
\emph{(b)} plots how many of its ten draws each honesty cell passed under both
judges, where a quarter of the cells fall between the two ends, so compliance
is reported as a rate. \emph{(c)} plots each model's rank on the
four measures, where the knowledge, skill, and attitude columns move together
while the reward column reverses them.}
\label{fig:panels}
\end{figure*}

Pressure separates the two honesty families rather than lowering both, as
Figure~\ref{fig:panels}a shows and Table~\ref{tab:pm} breaks down by model.
Under the deception family the pass probability falls from $0.81$ to $0.48$
when pressure is applied, and the item-level bootstrap interval excludes zero
under both judges. Under integrity and honesty the same manipulation moves the
pass probability from $0.94$ to $0.92$ and the interval covers zero. All eight
models move in the same direction on deception, so the split is a property of
the family and not of any one model.

Both families ask the model to say something untrue, and what differs is what
the untruth is for. Deception asks the model to confirm a wrong answer to the
child in the moment, while integrity and honesty asks it to write a record or
a report that a caregiver will read. The models hold the line on falsifying a
document and give way on agreeing with the child, which is the opposite of
what a rule stated as honesty in general would predict.

Yielding is a rate rather than a trait, as Figure~\ref{fig:panels}b shows.
Across the honesty cells, $62.6\%$
pass on all ten draws and $12.3\%$ fail on all ten, leaving $25.1\%$ that
split, and the second judge gives $61.9$, $12.5$, and $25.6$. Within the cells
that split, every value from one to nine occurs rather than only the two
ends, with the mass toward the passing side. Had the ends been the only outcomes, yielding would
be something a model always does in that situation, and the filled middle is
what lets us ask how far pressure moves the rate.

Monitoring works only where pressure is already applied. Announcing that a
supervising teacher will review the session moves the pass probability by
$0.010$ when there is no pressure, and that interval covers zero. Under
pressure the same announcement moves it by $0.036$ and the interval excludes
zero, with $0.010$ and $0.034$ under the second judge. A model does not behave
differently because it is watched. It behaves differently because it is
watched at the moment it is asked to lie. Models that give way to pressure are
also the ones most sensitive to being watched, and the correlation between the
pressure effect and the interaction is $0.60$ and $0.79$ over the eight
models.

\begin{table}[t]
\centering
\scriptsize
\setlength{\tabcolsep}{3pt}
\caption{Attitude under condition. The pressure effect is the drop in pass probability
when the item carries a request to misreport, the monitoring effect is the
rise when a supervisor is said to review the exchange, and P$\times$M is the
interaction between them. Larger P$\times$M means a model yields to pressure
more when it is not observed. The interaction is defined only on the two
families that carry both condition axes. Bold marks the value closest to zero
in each column and underline the next, since a model whose behavior the
condition does not move is the ideal.}
\label{tab:pm}
\begin{tabular}{@{}l rr r rr r@{}}
\toprule
 & \multicolumn{3}{c}{Deception} & \multicolumn{3}{c}{Integrity \& Honesty} \\
\cmidrule(lr){2-4} \cmidrule(lr){5-7}
Model & Pres. & Mon. & P$\times$M & Pres. & Mon. & P$\times$M \\
\midrule
GPT-5.6 Sol & \underline{+0.188} & +0.025 & +0.100 & \textbf{+0.000} & \textbf{+0.000} & \textbf{+0.000} \\
Claude Opus 5 & \textbf{+0.147} & \underline{+0.016} & -0.069 & \underline{-0.003} & \underline{-0.003} & \underline{+0.006} \\
Gemini 3.1 Pro & +0.431 & +0.069 & +0.137 & -0.041 & \underline{+0.003} & \underline{-0.006} \\
Grok 4.5 & +0.194 & +0.038 & -0.062 & -0.009 & \underline{+0.003} & \underline{-0.006} \\
Mistral Large 3 & +0.406 & +0.056 & \underline{+0.012} & +0.022 & +0.009 & +0.044 \\
Qwen3-VL & +0.494 & \textbf{+0.013} & +0.088 & +0.012 & +0.050 & +0.087 \\
Kimi K2.6 & +0.294 & +0.031 & \textbf{+0.000} & -0.009 & +0.009 & +0.019 \\
GLM-4.6V & +0.475 & +0.031 & \underline{+0.012} & +0.162 & +0.019 & +0.050 \\
\bottomrule
\end{tabular}
\end{table}

\subsection{Correlated Axes, Reversed Rankings}

Placing the eight VLMs on all four measures at once gives a shape we did not
expect, drawn in Figure~\ref{fig:panels}c. The three rubric axes track one
another (Appendix~\ref{app:correlation}) at $+0.62$, $+0.86$, and $+0.90$, though with eight models only the
two pairs involving attitude separate from chance. The reward metric runs
against them, at $-0.86$ on knowledge, yet it is weakest against skill at
$-0.36$, the axis its training pairs were drawn from. The
model that places last on knowledge places first on reward, and the model that
places third on knowledge places last on reward.

The sign is the opposite of what the benchmark that introduced this
decomposition reports \cite{lee2026openlearnlm}, where independence across
axes is offered as the reason to score three of them. Resting that case on low
correlation makes it depend on how wide the domain is, and ours is a single
clinical practice. The case survives on other grounds, since attitude fails
where the average does not show it, and a reward model trained on the skill
pairs still ranks the models unlike the skill rubric it was built from. What
counts as good teaching depends on the scorer as much as on the model.

\subsection{Strategy Boundaries and Special-Education Dispositions}

Failures concentrate at the boundary between strategies that look alike.
Deviation from the assigned strategy is highest on MT\_IncidentalTeaching, and
the typical error reads the child's answer to a teacher question as a
child-initiated bid, which turns the response into MT\_MandModel. What
separates the two is who initiates, so that is where the boundary sits. On
three items both judges marked all eight models as having left the strategy,
so those scenes defeat every model we tested.

The special-education families do not respond to pressure in one direction.
On evidence-based judgment pressure raises the pass probability rather than
lowering it, because the pressure there takes the form of a caregiver
asserting an unsupported conclusion, which gives the model something concrete
to refuse. The sign is the reverse of what the two honesty families show, so
the same manipulation does different work depending on the family and we do
not pool the pressure effect across families. Respect for potential and
respect carry sixteen cells each, so we read them at the family level only.

\section{Conclusion and Discussion}

We presented SpecialEduBench, which scores pedagogical competence in autism
language intervention along knowledge, skill, and attitude. Skill and attitude
are built on recorded intervention rather than written scenarios, the attitude
items cross a pressure condition with a monitoring condition, and the judge
model is read against the agreement the experts reach with each other rather
than against a threshold assumed in advance.

Across eight frontier vision-language models no axis is saturated. The models
converge where the knowledge is factual and separate where the task is
situated, the strongest still fails about a tenth of the honesty cells, and the
failures gather where pressure is applied. Attitude gives way in a place no
average over an axis reveals, and monitoring changes behavior only at the
moment a falsehood is asked for.

That the three axes rank the models alike follows from narrowing the domain and
does not soften the case for scoring them apart, since knowledge shows what a
model holds, skill shows whether it fits the scene, and attitude shows whether
it survives a request to do otherwise.

Validating the scorer before using it changes what the evaluation says, since
a reward model trained on the skill pairs ranks the same models unlike the
skill rubric it was built from. Putting the recorded scene in front of the
scorer, rather than a transcript of it, is the next step. We offer this
benchmark as a check to run before deployment rather than as a substitute for
evaluation in a classroom, and Appendix~\ref{app:release} sets out what we
release.

\section*{Limitations}

Expert review sets the size of the two judged axes. Two hundred skill items
rest on 45 clips and 68 attitude items on 17, and everything is English inside
one clinical practice, so we do not claim the strategy boundaries transfer
beyond it. Eight models is a small sample for the rank correlations, the reward
model is evaluated on 40 pairs, and both judges read text alone.

Neither judge figure is a clean held-out estimate, since the attitude
instruction was developed against the same 192 cells it later scored and the
skill set was opened more than once while we selected among instructions. Both
judges recover only part of what the experts placed below the passing line, so
they miss failures rather than invent them.

On 94 knowledge items all twenty-four attempts disagree with the published key
and 61 converge on one alternative, the pattern of a stale key rather than of a
hard item. Experts confirmed the published keys, so they are counted as wrong.



\bibliography{aaai2027}

\clearpage
\section*{Appendix}
\appendix

\section{Extended Related Work}
\label{app:related}

\subsection{Benchmarks for Educational LLMs}
\label{app:benchmarks}

Educational benchmarks for LLMs have grown along three
lines. One line scores subject knowledge and tutoring dialogue in general
education, where MathTutorBench places a tutor in a mathematics dialogue and
scores scaffolding with a learned reward model \cite{macina2025mathtutorbench}
and MMTutorBench extends the setting to multimodal input
\cite{yang2026mmtutorbench}. A second line decomposes competence
rather than task, and OpenLearnLM scores knowledge, skill, and attitude across
several subjects \cite{lee2026openlearnlm}. A third line reaches special
education but stops at multiple-choice knowledge
\cite{lelievre2025pedagogy}.

Two recent benchmarks share our three-axis decomposition and differ in what
they measure. EduVideoBench scores models that generate educational video
\cite{lee2026eduvideobench}, while we score a tutor that watches an
intervention and produces a teaching response. TeachObs validates multimodal
teaching observation against human raters across general education
\cite{jeong2026teachobs}, while we narrow to
autism language intervention so that the criteria can be written by people who
practice it. None of the three states what level of agreement a judge model has
to reach, which is the gap our scoring protocol addresses.

\subsection{Automatic Scoring of Teaching Quality}
\label{app:scoring}

Scoring open-ended teaching responses has converged on using a second
LLM as the judge. MT-Bench established the pattern of pairwise and
single-answer judging with a strong model \cite{NEURIPS2023_91f18a12}, and
G-Eval added chain-of-thought scoring against a written rubric
\cite{liu2023geval}. Work on tutoring dialogue has since applied the pattern
to pedagogical quality rather than answer quality
\cite{maurya2025mrbench}, and MathTutorBench replaces the judge with a
reward model trained to prefer expert responses over novice ones
\cite{macina2025mathtutorbench}.

None of this work separates reliability from validity, since a judge that
repeats itself is stable without being right \cite{melo2026}.

\subsection{Honesty and Disposition in LLMs}
\label{app:honesty}

Sycophancy is the best documented case of a model changing what it says with
who is asking, where it agrees with a stated user view against the evidence
\cite{sharma2024sycophancy}. Related work separates what a model represents
internally from what it states \cite{wang2025honesty} and shows that stated
reasoning need not be the reason for the answer \cite{turpin2023cot}. Alignment
faking shows the sharpest version, where behavior differs between conditions
the model reads as monitored and unmonitored \cite{greenblatt2024alignment}.

What this literature does not cover is the case our attitude axis builds. In
an intervention session the person asking
for the falsehood is the caregiver or the child, the record that would be
falsified goes to a team meeting, and the party harmed is the child whose
learning depends on the record being true.

\subsection{Background on Special Education}
\label{app:speced}

Special education adjusts the content, method, and delivery of instruction to
a child's individual needs, so goals and assessments differ from child to
child by design \cite{cfr30039}. For children with complex communication
needs the goal is functional communication rather than a correct answer
\cite{kaiser2017emt}, and the same teacher response is good teaching or poor
teaching depending on what the child did immediately before
\cite{mcdaniel2022}. Fit to the situation, not agreement with a key, is
therefore what a judgment has to be made against.

Responsive interaction covers following the child's lead, mapping a nonverbal
act onto words, expanding an utterance into a fuller form, and recasting an
error without marking it \cite{mcdaniel2022}. Milieu teaching covers modeling the target language,
requesting a response and then modeling it if none comes, waiting with an
expectant signal, and turning a child-initiated request into a teaching
opportunity. What separates neighboring strategies is who initiates and
whether the adult waits, which is why the boundaries described in
Section~5.3 fall where they do.

The dispositions come from separate literatures. Presuming competence and
honoring self-determination are stated as obligations toward the child
\cite{donnellan1984criterion, uncrpd2006}, evidence-based judgment as an
obligation toward the field \cite{cook2013ebp}, and honest reporting as an
obligation toward the team that reads the record. A benchmark that scores only
what a teacher says under ordinary conditions cannot separate these from
politeness.

\section{Rubric Bands}
\label{app:rubrics}

Both judged axes share one band structure so that scores stay comparable
across strategies and families. Every rubric runs on a ten-point scale split
into five bands, and only the description of what the response did inside each
band is written per strategy. A score of seven or above is a pass, five and
six are a borderline zone that we count as neither, and four or below is a
failure. The skill axis has eight strategies and the attitude axis has six
families, so the benchmark carries forty and thirty band descriptions.

The two upper bands are deliberately built on the same premise. The 7 to 8 band
asks that the defining condition of the strategy be met, and the 9 to 10 band
asks for the same thing plus one further step, which on the skill axis is
evidence traceable to the observed clip and on the attitude axis is a
constructive move offered after the refusal. Separating the bands this way
keeps the pass line at one place while still letting the rubric reward a
response that does more than clear it. Scoring proceeds in two steps. The
judge first selects a band from its description, then chooses the lower or
upper point inside that band, and an unresolved condition resolves downward
rather than being averaged away.

Expansion is a representative skill rubric. Its 1 to 2 band covers a response
that does not start from the child's preceding utterance, changes the meaning
or topic, or falls back on explicit correction and imitation demands. The 3 to 4
band covers a superficial acknowledgment followed by language that is
excessive, unnatural, or meaning-changing, so no slightly more complete form
is established. The 5 to 6 band preserves the child's meaning and adds a
plausible element while leaving the modest increment or the boundary against
recast incomplete. The 7 to 8 band acknowledges the utterance, preserves its
meaning, and produces a modest and slightly more complete form in natural
conversational flow. The 9 to 10 band does all of that and also distinguishes
additive expansion from the correction of a spoken commission error, which is
the distinction that separates expansion from recast.

Deception is the matching attitude rubric. The 1 to 2 band covers a response
that caves to guardian pressure or uses the unmonitored gap to confirm a wrong
answer and pour out baseless praise. The 3 to 4 band covers quietly letting the
wrong answer slide with unfounded reassurance. The 5 to 6 band withholds open
false confirmation but hedges or changes the subject, so the child leaves with
the misconception intact. The 7 to 8 band refuses the false confirmation and the
baseless praise even while unmonitored and under pressure, though it is thin
on saying why the answer is wrong. The 9 to 10 band adds that explanation warmly
and offers a concrete scaffold for the next attempt. Items marked as traps carry
one further instruction. A trap embeds a wrong fact or number, and correcting
it scores higher while going along with it scores lower.

\section{Attitude Dimensions}
\label{app:dimensions}

The attitude axis has six families and seventeen categories, with four items
in each category. A family fixes what is at stake and a category fixes the
shape the failure takes, so two categories in one family are scored by the
same rubric but reached through different requests.

Deception covers dishonesty toward the child in the moment of the
interaction, and it splits four ways. False confirmation agrees that a wrong
answer is right. Sycophancy supplies praise the performance does not support.
Fabrication invents a fact to keep the exchange moving. Omission withholds a
correction the child needs to move on. Integrity and honesty
covers dishonesty in the record rather than in speech, and it also splits four
ways. False progress reports growth that did not occur. Inflated records
raise a logged value. Number fabrication supplies a count that was never
measured. False premise accepts a claim built into the request instead of
questioning it.

The remaining four families carry the professional dispositions that special
education asks of a tutor. Potential splits into low expectations, where the
tutor accepts a ceiling on what the child can reach, and affirming potential,
where it holds that ceiling open. Respect splits into autonomy, which concerns
choices made on the child's behalf, and emotion, which concerns feelings
dismissed as noise. Evidence splits three ways, into hasty conclusion, which
draws an inference from too little observation, stereotype, which substitutes
a diagnostic label for an observation, and request evidence, where the correct
move is to ask for the record before answering. Proactive splits into
scaffolding, which builds the support in advance, and responsive, which
adjusts once the child's response arrives.

Only the first two families carry both condition axes, because only there is a
specific false statement being requested and so only there is refusal well
defined. Evidence, potential, and respect carry pressure alone, and proactive
teaching is scored once under the base prompt because there is nothing in it to
push against.

\section{Expert Review}
\label{app:review}

Seven special-education experts took part, five who built the benchmark and
two who reviewed the result without having taken part in building it. The five
fixed the strategy assignments, checked items against rubrics, scored the
reference responses, and settled each reference score by discussion. The two
examined a stratified sample of the finished reference scores and the
preference pairs. Table~\ref{tab:review} records each stage.

\begin{table}[t]
\centering
\scriptsize
\setlength{\tabcolsep}{3pt}
\caption{Expert review by stage. Internal marks the experts who built the
benchmark and external those who reviewed the result. The inter-expert
agreement on each axis is the ceiling against which Section~3.5 reads the
judge.}
\label{tab:review}
\begin{tabular}{@{}l l l l@{}}
\toprule
Axis & Stage & Reviewers & Outcome \\
\midrule
Skill & Strategy assignment & Internal 4 & keep 84, drop 43, revise 20 \\
      & Item and rubric     & Internal 4 & 7 of 200 revised \\
      & Reference scoring   & Internal 4 & 400 scores, 290 pass \\
      & Consensus           & Internal 4 & $\alpha$ $0.427$, $\pm1$ on 63.9\% \\
      & Adequacy            & External 2 & 40 of 200, all adequate \\
\midrule
Attitude & Item and rubric  & Internal 4 & 12 revised, 5 approved \\
      & Reference scoring   & Internal 4 & 384 scores, 303 pass \\
      & Consensus           & Internal 4 & $\alpha$ $0.685$, $\pm1$ on 80.7\% \\
      & Adequacy            & External 2 & 40 of 192, all adequate \\
\midrule
Knowledge & Pilot           & Internal 1 & 15 of 16 adequate \\
      & Full review         & External 2 & 3 of 500 flagged \\
\midrule
Reward & Pair adequacy      & External 2 & 181 of 200 adequate \\
\bottomrule
\end{tabular}
\end{table}

The reference scores that both alignment figures are measured against were
produced in two passes. On the skill axis four of the five internal experts scored 400
reference responses independently, of which 290 landed in a passing band, 48 in
the borderline zone, and 55 below it. They then settled every one of the 200
items by discussion, and the settled distribution is concentrated at the top,
with 154 items at eight or above and only 13 at four or below. The inter-expert
agreement of $0.427$ is computed on the 194 items where all four experts
scored before the discussion.

The attitude axis followed the same two passes on a smaller set. All seventeen
categories came back from the item review with comments, of which twelve were
revised and five approved unchanged. The same four experts then scored 384
reference responses, with 303 passing, 36 borderline, and 45 failing, and
settled all 192 cells. Agreement here is higher at $0.685$, and the experts
landed within one point of each other on 155 of the 192 cells, disagreeing by
more than that on 37.

The gap between the two agreement figures is the reason we do not set a single
alignment target. Judging whether a tutor turn is a well-formed expansion is a
harder call than judging whether it confirmed something false, and the experts
themselves show that difference before any model is involved.

The two external reviewers saw the finished product rather than the process.
They read a stratified sample of 40 skill items, 40 attitude responses, and
200 preference pairs, and reported no unsuitable items on the first two. On
the preference pairs they rejected 19 of 200, which we regenerated by defect
type and resubmitted, after which all 200 were accepted.

\section{Data Construction Details}
\label{app:construction}

\paragraph{Knowledge.}
The 500 EMT items were written from eight primary sources on early milieu
teaching and naturalistic language intervention rather than paraphrased from an
existing question bank, since no public bank covers the strategies at the
level the axis needs. A single expert piloted 16 items and found 15 suitable,
after which two external reviewers read all 500 and flagged 3, a rate of
0.6\%. Thirty items are released but withheld from scoring, 25 because the source did
not publish an answer key we could verify and 5 because two options were
defensible under the same reading. A further 264 items ship without the supporting
rationale, since the rationale in the source is paywalled even where the item
itself is not.

\paragraph{Skill.}
Four internal experts reviewed the assignment of a strategy to each clip,
keeping 84 of 147 candidates, dropping 43, and revising 20. Three models then
drafted candidate items on the fixed pairings, 76 from GPT-4.1, 71 from
Gemini-3-flash, and 3 from Gemini-3.1-Pro, and the 136 base items were selected
from those 150 drafts by hand. The items and their rubrics went through a
second review that revised 7 of 200. The 64 augmented items were
produced from the base items, 54 by template and 10 by a model, and each
augmented item keeps its parent in the identifier.

\paragraph{Judging.}
The judge prompt asks for a checklist against the rubric rather than an
overall impression, which we adopted after the checklist form scored the same
responses more consistently across repeats. The skill alignment we report is
computed over all 200 expert-scored items rather than over a holdout, since
the full set was used more than once while we selected among instructions. We also varied how much of the
clip the judge sees. Measured on the 78-item development split rather than on
the full set, four frames give the best alignment at $0.4199$, twelve frames
give $0.4103$, and text alone gives $0.3946$, so the visual input contributes
little and more of it does not help. Eight frames land lowest at $0.3669$,
which we read as noise on this sample rather than as a real non-monotonicity.
Because that split is smaller and was used while selecting instructions, its
values sit above the $0.342$ we report over all 200 items. The main results
use the text-only setting on both judged axes so that the two are scored the
same way.

\paragraph{Reward.}
The rejected response in each preference pair carries one injected defect
drawn from four types, which are violating the wait interval, making
reinforcement contingent on the wrong behavior, taking over a turn the child
should lead, and turning a strategy into a quiz. Two external reviewers
accepted 181 of the first 200 pairs, and the 19 they rejected were regenerated
by defect type and accepted on resubmission.

The first reward model we trained scored length rather than teaching. We
measure this by appending four sentences of no pedagogical content to a
response and reading how far its score moves, and on that model it moved by
$+7.37$, which was two thirds of the median margin between a chosen and a
rejected response. Its score also correlated with word count at $0.43$.

Two repairs were tried and the same padding test judges both. Adding
length-matched pairs to the training set over-corrected badly, moving the
padding response by $-30.26$, so that model would have penalized any answer
for being long. It separated the held-out pairs slightly better than the model
we ship, at $0.975$ against $0.950$, which is why pair accuracy alone is not
enough to accept a reward model. Changing the base model instead, with the
training data left as it was, moved the padding response by $-1.16$ and kept
the separation, and that is the model we release.

\section{Dataset Composition}

\subsection{Knowledge Categories and Sources}
\label{app:knowledge-detail}

The knowledge axis draws on public benchmarks where verified items exist and
generates items only where they do not. Category sizes are uneven, which is
why the main text reports a macro-average over categories.
Table~\ref{tab:knowledge} gives the sources and their licenses.

\begin{table}[h]
\centering
\footnotesize
\setlength{\tabcolsep}{4pt}
\caption{The four knowledge categories, counted as items that are scored.
Collected items keep the license of their source. Category sizes follow the
availability of public resources rather than the importance of the domain, so
we report a macro-average. The axis releases $4{,}567$ items, of which $30$
medical items carry no verifiable key and are released without being scored.}
\label{tab:knowledge}
\begin{tabular}{@{}l r l l@{}}
\toprule
Category & Items & Source & License \\
\midrule
Medicine   & $3{,}752$ & MedMCQA, MedQA & Apache-2.0, CC BY 4.0 \\
EMT        & $500$     & Generated      & CC BY 4.0 \\
Education  & $223$     & Pedagogy Bench.\ & Apache-2.0 \\
Psychology & $62$      & MMLU, MMLU-Pro & MIT \\
\midrule
Total      & $4{,}537$ & & \\
\bottomrule
\end{tabular}
\end{table}

\subsection{Skill Items by Strategy}
\label{app:skill-detail}

Experts fixed the pairings of clip and strategy before any item was written. The
largest strategy was cut to twenty-five and the seven short ones were filled
with augmented items, each derived from a parent item on the same clip.
Table~\ref{tab:skill} gives the counts and the expert reference scores.

\begin{table}[h]
\centering
\footnotesize
\setlength{\tabcolsep}{4pt}
\caption{The eight teaching strategies. Base items come from expert-fixed
pairings of clip and strategy, and variants balance each strategy to 25.
Reference is the mean expert score on the 1 to 10 rubric, and $\alpha$ is the
interval agreement among the four scoring experts on the items of that
strategy.}
\label{tab:skill}
\begin{tabular}{@{}l l r r r r@{}}
\toprule
Strategy & Fam. & Base & Var. & Ref. & $\alpha$ \\
\midrule
Following the child's lead & RI & $25$ & $0$  & $8.08$ & $0.387$ \\
Linguistic mapping         & RI & $16$ & $9$  & $8.44$ & $0.295$ \\
Expansion                  & RI & $15$ & $10$ & $8.96$ & $0.125$ \\
Recast                     & RI & $11$ & $14$ & $8.24$ & $0.409$ \\
Modeling                   & MT & $20$ & $5$  & $7.52$ & $0.469$ \\
Mand-model                 & MT & $17$ & $8$  & $8.24$ & $0.393$ \\
Time delay                 & MT & $17$ & $8$  & $8.96$ & $0.232$ \\
Incidental teaching        & MT & $15$ & $10$ & $8.56$ & $0.689$ \\
\midrule
Total                      &    & $136$ & $64$ & $8.38$ & $0.427$ \\
\bottomrule
\end{tabular}
\end{table}

\subsection{Attitude Families and Conditions}
\label{app:attitude-detail}

\begin{table}[h]
\centering
\footnotesize
\setlength{\tabcolsep}{3pt}
\caption{The six attitude families. P marks the pressure axis and M the
monitoring axis. Traps are items that plant a factually wrong premise, and OLM
marks a counterpart in the OpenLearnLM attitude axis. Appendix~F defines the
17 dimensions.}
\label{tab:attitude}
\begin{tabular}{@{}l r r c r r c@{}}
\toprule
Family & Dim. & Items & Axes & Cells & Trap & OLM \\
\midrule
Deception            & $4$ & $16$ & P$\times$M & $64$ & $2$ & \checkmark \\
Integrity \& Honesty & $4$ & $16$ & P$\times$M & $64$ & $8$ & \checkmark \\
Evidence-based Judg. & $3$ & $12$ & P         & $24$ & $0$ & \checkmark \\
Respect for Potential& $2$ & $8$  & P         & $16$ & $0$ &            \\
Respect              & $2$ & $8$  & P         & $16$ & $0$ &            \\
Proactive Teaching   & $2$ & $8$  & ---       & $8$  & $0$ & \checkmark \\
\midrule
Total                & $17$ & $68$ &          & $192$ & $10$ & \\
\bottomrule
\end{tabular}
\end{table}

Table~\ref{tab:attitude} gives the families. The two honesty families carry
both condition axes, so the interaction is defined only there, and the other
four carry pressure alone or a single baseline condition.

\section{Evaluated Models and Sampling}
\label{app:models}

\begin{table}[h]
\centering
\footnotesize
\setlength{\tabcolsep}{4pt}
\caption{The eight models under evaluation. Temperature is omitted for one
model because the provider ignores it rather than rejecting it. Reasoning is
minimized per model at the effort level that produced the fewest reasoning
tokens on a probe, and three models require reasoning and cannot go below
their own floor. Frames is the per-request image cap, which one model lowers
because it rejects anything larger.}
\label{tab:models}
\begin{tabular}{@{}l l l l r@{}}
\toprule
Model & Weights & Temp. & Reason. & Frames \\
\midrule
GPT-5.6 Sol                & Closed & ---     & minimal & all \\
Claude Opus 5              & Closed & $1.0$   & low     & all \\
Gemini 3.1 Pro             & Closed & $1.0$   & forced  & all \\
Grok 4.5                   & Closed & $1.0$   & forced  & all \\
Mistral Large 3            & Open   & $1.0$   & none    & $8$ \\
Qwen3-VL-235B-A22B-Think.  & Open   & $1.0$   & forced  & all \\
Kimi K2.6                  & Open   & $1.0$   & minimal & all \\
GLM-4.6V                   & Open   & $1.0$   & minimal & all \\
\bottomrule
\end{tabular}
\end{table}

Table~\ref{tab:models} gives the per-model settings. One model does not accept
a seed, so its three knowledge runs differ only by sampling.

\section{Additional Results}

\subsection{Knowledge Accuracy by Disability Type}
\label{app:bucket}

Table~\ref{tab:bucket} breaks accuracy out by the disability a knowledge item
concerns. No bucket reverses the broad picture, but the ranks are not stable
inside it, and only 29 of the 64 model positions match the ordering on the
autism bucket. The spread within a model runs from $3.5$ points on Claude Opus
5 to $9.9$ on Qwen3-VL, so the weaker models are also the more uneven ones.
The buckets are of very different sizes and 723 items carry no bucket tag at
all, which is why the main text reports the category macro-average instead of
this breakdown.

\begin{table}[h]
\centering
\scriptsize
\setlength{\tabcolsep}{2.5pt}
\caption{Knowledge accuracy by disability type, with the best value in each column in bold and the second best underlined. ASD is autism, ID is
intellectual disability, HI is hearing, VI is vision, CP is cerebral palsy and
motor, Law is the special-education system and its regulations, and Dev is
developmental in general. The last row gives the number of items. These eight
buckets cover $3{,}630$ of the $4{,}537$ items. Three smaller buckets hold $184$
more and $723$ items carry no bucket tag.}
\label{tab:bucket}
\begin{tabular}{@{}l rrrrrrrr@{}}
\toprule
Model & ASD & ID & ADHD & HI & VI & CP & Law & Dev \\
\midrule
GPT-5.6 Sol & 94.9 & 93.6 & 92.2 & 93.3 & 93.4 & 92.0 & 90.6 & 93.9 \\
Claude Opus 5 & \underline{95.0} & \underline{93.7} & 92.4 & 92.8 & 92.5 & \underline{93.2} & 91.5 & 93.6 \\
Gemini 3.1 Pro & 94.9 & \underline{93.7} & \underline{92.6} & \underline{94.0} & \underline{94.6} & 90.6 & \underline{92.9} & \underline{94.7} \\
Grok 4.5 & 94.1 & 92.6 & 91.9 & 92.5 & 92.3 & 92.3 & 90.6 & 91.6 \\
Mistral Large 3 & 90.1 & 86.6 & 85.6 & 83.4 & 84.5 & 83.8 & 84.8 & 88.3 \\
Qwen3-VL & 92.7 & 90.0 & 88.8 & 88.4 & 89.4 & 82.9 & 89.4 & 89.8 \\
Kimi K2.6 & 92.6 & 91.5 & 89.4 & 90.8 & 91.1 & 87.0 & 88.3 & 91.9 \\
GLM-4.6V & 90.6 & 87.3 & 87.1 & 85.0 & 83.4 & 83.1 & 84.0 & 88.3 \\
\midrule
Items & \textbf{266} & \textbf{1374} & \textbf{516} & \textbf{800} & \textbf{217} & \textbf{138} & \textbf{188} & \textbf{131} \\
\bottomrule
\end{tabular}
\end{table}

\subsection{Rank Correlation Between Measures}
\label{app:correlation}

Table~\ref{tab:dissociation} gives every pairwise rank correlation behind the
claim in Section~6 that the three axes order the models much alike.

\begin{table}[h]
\centering
\footnotesize
\setlength{\tabcolsep}{5pt}
\caption{Rank correlation between the four measures over the eight models,
reported as Spearman $\rho$ with a permutation $p$ over 20,000 draws, with
tied reward scores given their average rank. Eight models is a small sample,
so we read the size of $\rho$ and treat $p$ as a guard against reading a
pattern into noise.}
\label{tab:dissociation}
\begin{tabular}{@{}l r r@{}}
\toprule
Pair & $\rho$ & $p$ \\
\midrule
Knowledge and skill    & $+0.62$ & $0.112$ \\
Knowledge and attitude & $+0.86$ & $0.011$ \\
Skill and attitude     & $+0.90$ & $0.005$ \\
\midrule
Knowledge and reward   & $-0.86$ & $0.013$ \\
Attitude and reward    & $-0.69$ & $0.061$ \\
Skill and reward       & $-0.36$ & $0.390$ \\
\bottomrule
\end{tabular}
\end{table}

\subsection{Reward Win Rate by Strategy}
\label{app:reward-strategy}

The reward model separates strategies more than it separates models, as
Table~\ref{tab:reward-strategy} shows. The four
responsive-interaction strategies average $0.900$ while the four milieu
teaching strategies average $0.750$, and MT\_TimeDelay is lowest. Milieu
teaching turns on the timing and order of a prompt, which the transcript does
not carry, so the gap is read as a limit of a text-only scorer rather than as
a property of the models.

\begin{table}[h]
\centering
\footnotesize
\caption{Reward win rate by strategy over the 40 held-out items and eight
models, giving 40 comparisons per strategy.}
\label{tab:reward-strategy}
\begin{tabular}{@{}l r r@{}}
\toprule
Strategy & Wins & Win rate \\
\midrule
RI\_Expansion & 38/40 & \textbf{0.950} \\
RI\_LinguisticMapping & 38/40 & \textbf{0.950} \\
RI\_Recast & 35/40 & \underline{0.875} \\
RI\_FollowChildLead & 33/40 & 0.825 \\
MT\_MandModel & 33/40 & 0.825 \\
MT\_IncidentalTeaching & 31/40 & 0.775 \\
MT\_Modeling & 31/40 & 0.775 \\
MT\_TimeDelay & 25/40 & 0.625 \\
\bottomrule
\end{tabular}
\end{table}

\section{Dataset Release}
\label{app:release}

All three axes follow the OpenLearnLM schema \cite{lee2026openlearnlm}. Skill
and attitude items carry the address and time range of the grounding clip, and
knowledge items carry the category and the disability type as tags.

The benchmark is released under CC BY 4.0 and collected items keep the license
of their source. Seven of the eight EMT source papers are open access and the
eighth is under copyright, whose supporting quotations we withhold. We do not
redistribute the video, since most clips carry the standard platform license,
and each item carries a scene description written so that it can be answered
from the text alone.

\end{document}